# Exploring Thermography Technology: A Comprehensive Facial Dataset for Face Detection, Recognition, and Emotion

## Authors
Mohamed Fawzi Abdelshafie Abuhussein[1&2&*], Ashraf Darwish[2&3], Aboul Ella Hassanien[2&4]
**Affiliations**
[1]Department of Agricultural Constructions Engineering and Environmental Control, Faculty of Agricultural Engineering, Al-Azhar University, Cairo 11751, Egypt
[2]Scientific Research School of Egypt (SRSEG), Cairo, Egypt
[3]Faculty of Science, Helwan University, Helwan, Cairo, Egypt
[4]Faculty of Computers and Artificial Intelligence, Cairo University, Cairo, Egypt
corresponding author(s): Mohamed Fawzi Abdelshafie Abuhussein (mohamedfawzi8898@gmail.com)

## Abstract
This dataset includes 6823 thermal images captured using a UNI-T UTi165A camera for face detection, recognition, and emotion analysis. It consists of 2485 facial recognition images depicting emotions (happy, sad, angry, natural, surprised), 2054 images for face recognition, and 2284 images for face detection. The dataset covers various conditions, color palettes, shooting angles, and zoom levels, with a temperature range of -10°C to 400°C and a resolution of 19,200 pixels. It serves as a valuable resource for advancing thermal imaging technology, aiding in algorithm development, and benchmarking for facial recognition across different palettes. Additionally, it contributes to facial motion recognition, fostering interdisciplinary collaboration in computer vision, psychology, and neuroscience. The dataset promotes transparency in thermal face detection and recognition research, with applications in security, healthcare, and human-computer interaction.

## Background & Summary
The motivation behind compiling this dataset stems from the increasing interest in thermal imaging technology and its applications, particularly in the field of computer vision. Thermal imaging offers unique advantages, such as the ability to capture temperature variations and operate in low-light or adverse environmental conditions. We aimed to create a comprehensive dataset of thermal face images captured using different color palettes, enabling researchers to develop and evaluate algorithms for thermal face detection, recognition, and facial motion analysis. By providing diverse and standardized data, we sought to contribute to the advancement of research in thermal imaging technology and its interdisciplinary applications in fields such as biometrics, security, healthcare, and non-verbal communication analysis.

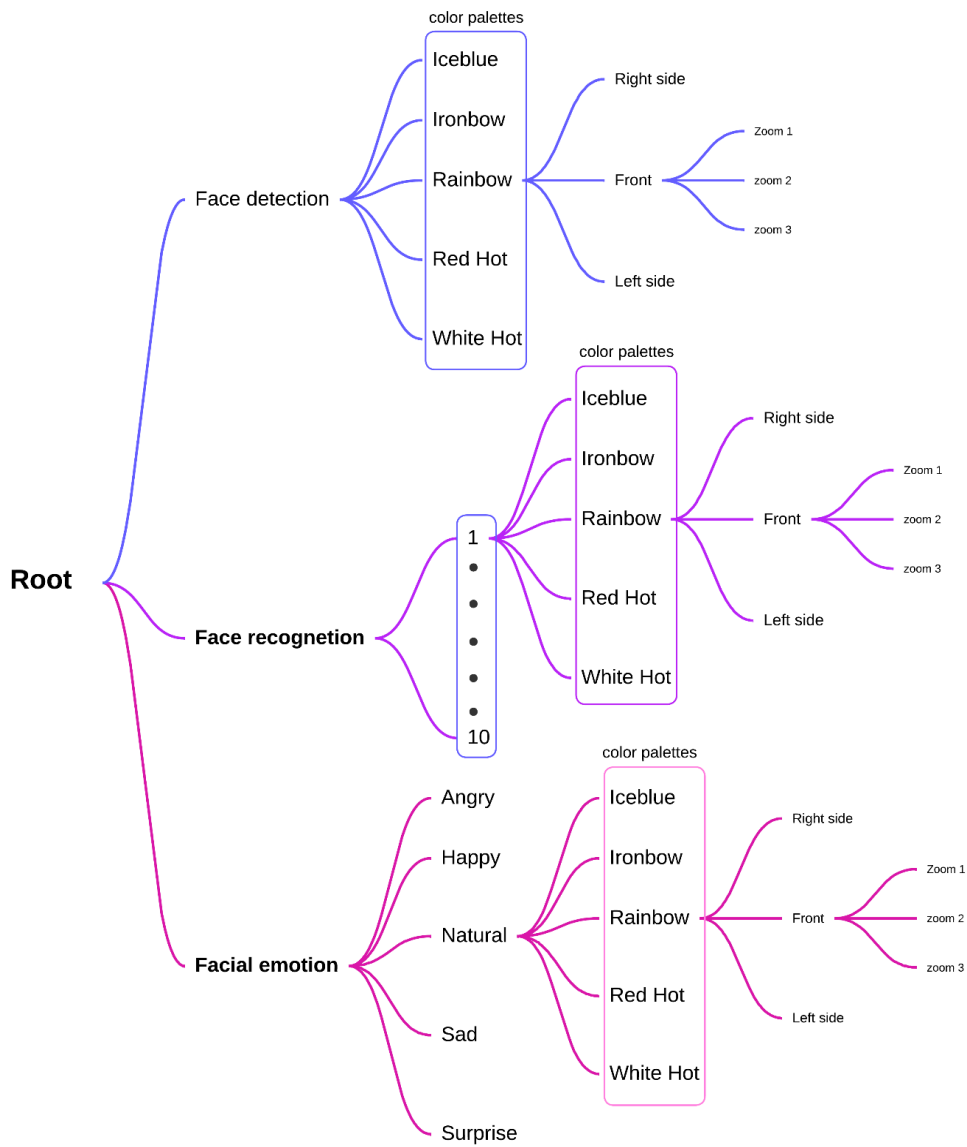

**Fig. 1.** Data structure of the repository

**VALUE OF THE DATA**
- **Valuable dataset for advancing thermal imaging technology:** The diverse collection of thermal face images, captured with different color palettes, facilitates research on image processing algorithms, pattern recognition, and machine learning techniques tailored to thermal imaging.
- **Facilitates benchmarking and comparison:** Researchers can utilize these data to benchmark and compare the effectiveness of thermal face detection and recognition algorithms across various color palettes. This aids in developing robust methods for real-world applications.
- **Contributes to facial motion recognition studies:** With thermal images suitable for facial motion recognition, this dataset offers insights into non-verbal communication analysis and emotion detection using thermal cues, benefiting research in psychology and neuroscience.
- **Encourages interdisciplinary collaboration**: These data foster collaboration among researchers in computer vision, image processing, psychology, and neuroscience, leading to innovative applications in biometrics, security, and healthcare.

- **Enhances accessibility and reproducibility:** Open sharing of these data promotes transparency and scientific advancement by providing access to resources for studies related to thermal face detection and recognition.

**Research Directions and Applications of Thermal Imaging Facial Dataset**

The development of a dataset about Thermal Imaging Technology for Face Detection, Recognition, and Facial Emotion Analysis has the potential to significantly expand the scope of study prospects and practical implementations for academics and researchers. The dataset has the potential to be exploited in various sectors, research topics, and applications.

- **Facial Detection and Recognition**
  
  The dataset enables researchers to explore the effectiveness of thermal imaging compared to visible light for face detection and recognition, as well as develop algorithms resilient to adverse conditions [1,2]
  a) To examine the comparative efficacy of thermal imaging about visible light for face detection and recognition.
  b) To design and assess algorithms that can effectively detect and recognize faces in situations with limited lighting or unfavorable environmental circumstances, utilizing thermal imaging data.
  c) To investigate the resilience of thermal imaging-based facial recognition systems in the face of fluctuations in facial expressions, occlusions, and lighting conditions.

- **Facial Emotion Analysis**
  
  R, Researchers can utilize the dataset to delve into thermal-based emotion analysis and its applications in healthcare, biometrics, security, and human-computer interaction [3,4]
  a) To conduct thermal image analysis to identify and classify face expressions linked to different emotions.
  b) To create emotion identification algorithms by utilizing thermal face data and evaluating their efficacy compared to visible light-based approaches.
  c) To examine the influence of temperature fluctuations and physiological reactions on emotion recognition systems that rely on thermal data.

- **Healthcare and Biometrics**
  
  Thermal face data has the potential to be utilized in health monitoring applications, specifically in the diagnosis of fever, as well as in biometric authentication systems. This is particularly relevant in situations where approaches based on visible light may exhibit lower reliability [5]
  a) The application of thermal facial data in health monitoring applications involves the detection of physiological irregularities, such as fever.
  b) Investigating the possibilities of thermal imaging in biometric authentication systems, with a specific focus on situations where visible light-based approaches may exhibit reduced reliability, such as in low-light conditions or when faces are heavily obscured.

- **Security and Surveillance**
  
  The dataset provided facilitates the implementation of thermal-based face identification and recognition systems for surveillance and security applications, particularly in situations characterized by low-light or nighttime settings [6]
  a) Deploying thermal-based facial detection and identification systems for surveillance and security purposes, particularly in environments with limited illumination or during nighttime hours.
  b) Examining the efficacy of thermal imaging in the detection of concealed or masked faces, a task that might pose difficulties for systems relying solely on visible light.

- **Human-Computer Interaction (HCI)**

Integrating thermal-based facial recognition and emotion analysis into HCI systems allows for the development of personalized user interfaces and emotion-aware computing devices, expanding the possibilities of human-computer interaction [7]
- a) The incorporation of thermal-based facial recognition and emotion analysis into HCI systems, such as personalized user interfaces or emotion-aware computing devices, is under consideration.
- b) Investigating innovative uses of thermal imaging technologies in HCI, such as utilizing face motions recognized through thermal data to provide hands-free control.

- **Cross-Modal Analysis**
Researchers can enhance the precision and resilience of face identification, recognition, and emotion analysis systems by integrating thermal facial data with other modalities, such as visible light images or auditory signals, through multimodal analysis [8]
  - a) The integration of thermal facial data with additional modalities, such as visible light images, depth data, or audio signals, is employed to enhance face detection, recognition, and emotion analysis through multimodal analysis.
  - b) Exploring fusion strategies that can effectively utilize the complementary information offered by several modalities to enhance the robustness and accuracy of the obtained findings.

- **Ethical and Privacy Considerations**
It is imperative to prioritize the examination of ethical considerations about data privacy, biases, and the responsible utilization of thermal imaging technology. This is essential to guarantee fair and ethical implementation of facial analysis systems across many societal settings
  - a) Examining the ethical considerations associated with acquiring and utilizing thermal facial data, with a particular focus on safeguarding data privacy and upholding personal rights.
  - b) Investigating the potential biases and limits associated with thermal imaging technology in the context of facial analysis, with a specific focus on its applicability to varied populations and cultural backgrounds.

The Thermal Imaging Technology dataset is a valuable resource for scholars and researchers in various fields, including face detection, recognition, and facial emotion analysis. Researchers have the opportunity to investigate a range of research avenues and applications by utilizing thermal imaging data. These include but are not limited to facial identification and recognition, emotion analysis, healthcare monitoring, security and surveillance, human-computer interaction, and cross-modal analysis. This dataset provides opportunities to examine the efficacy of thermal imaging in difficult circumstances, such as low-light settings, while also taking into account ethical and privacy concerns. Researchers have the potential to make significant contributions to the fields of computer vision, biometrics, and personalized computing systems by incorporating thermal face data with other modalities and creating resilient algorithms.

## Methods
**SPECIFICATIONS TABLE**
This section lists and summarizes the data collection procedure and format of the thermal imaging dataset.

| Subject | Human-Computer Interaction, Computer Vision and Pattern Recognition, Artificial Intelligence |
|---|---|
| Specific subject area | Face Detection, Recognition, and Emotion Analysis through Thermal Imaging: Utilizing heat signatures to identify and track human faces, recognize individuals, and analysis facial expressions in thermal data. |

| Type of data | Image (.BMP) |
|---|---|
| Data collection | Thermal face images from diverse individuals were captured using a UNI-T UTi165A camera. The camera features a temperature range of -10°C to 400°C with a measurement resolution of 0.1°C and an IR resolution of 19,200 pixels. Various color palettes were employed during image capture. |
| Data source location | The thermal imaging dataset was collected and stored by the Scientific Research Group in Egypt (SRGE), located in Giza, Egypt. This institution conducted the data collection, processing, and storage related to the thermal images used in the dataset. |
| Data accessibility | Repository name: Comprehensive Facial Thermal Dataset<br>Direct URL to data:<br>https://data.mendeley.com/preview/8885sc9p4z?a=07512cc6-5b3e-4c82-a493-5d4e87534de6 [11] |
| Related research article | Palani Thanaraj Krishnan. (2023). Thermal and Visible Images. IEEE Dataport. https://dx.doi.org/10.21227/309c-5b85 [12] |

**Thermal images data collection**

Thermal images in this dataset were captured using a UNI-T UTi165A thermal imaging camera, produced by Uni-Trend Technology Company Limited in Dongguan City, Guangdong Province, China. The camera features a temperature range of -10°C to 400°C, with a measurement resolution of 0.1°C. Its infrared (IR) resolution is 19,200 pixels [13]. For detailed specifications of the thermal camera, please refer to **Table 1.**

**Table 1** The detailed specifications of the thermal camera

| | |
|---|---|
| **IR Resolution (pixels)** | 160×120 |
| **Display** | 2.8" TFT LCD (320×240) |
| **Thermal Camera sensitivity (NETD)** | ≤50mk |
| **Spatial resolution (IFOV)** | 6.1mrad |
| **Camera Frame rate (Hz)** | ≤9 |
| **Spectral range (μm)** | 8~14 |
| **Temperature measurement range (°C)** | -10~400 |
| **Accuracy** | ±2°C or ±2% whichever is greater |
| **Emissivity** | 0.01~1.00 adjustable (0.95 default) |
| **Color palettes** | (White hot, Red hot, Ice blue, Rainbow, Ironbow) |
| **Image/Video storage format** | BMP |
| **UTi165A Certificates** | CE, FCC, RoHS |
| **Product weight (g)** | 490 |
| **Product size (mm)** | 236×75.5×86 |

**Color palettes.**

In thermal imaging, "color palettes" refer to a set of color schemes used to represent different temperature levels with specific colors as illustrated in **Fig.8**. These palettes provide a visual means to interpret thermal data, where different temperature levels are assigned distinct colors symmetrically across the image [14]. A variety of color palettes are used in our dataset, including ICEBLUE, IRONBOW, RAINBOW, Red Hot, and White Hot, as illustrated in **Fig. 9**. These palettes vary in how they assign colors to different temperature levels, offering multiple options for representing thermal data in different ways according to user needs and various applications.

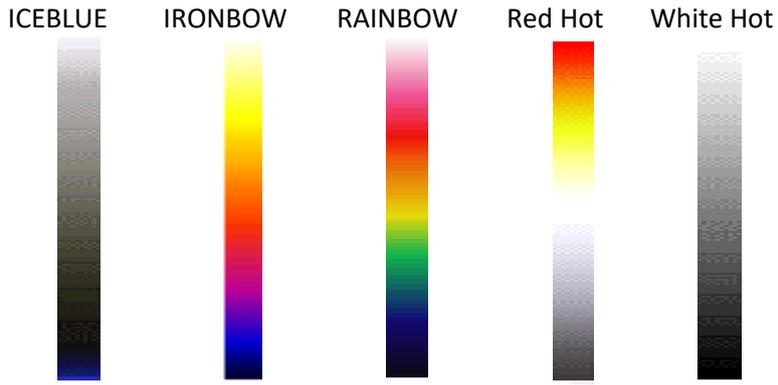

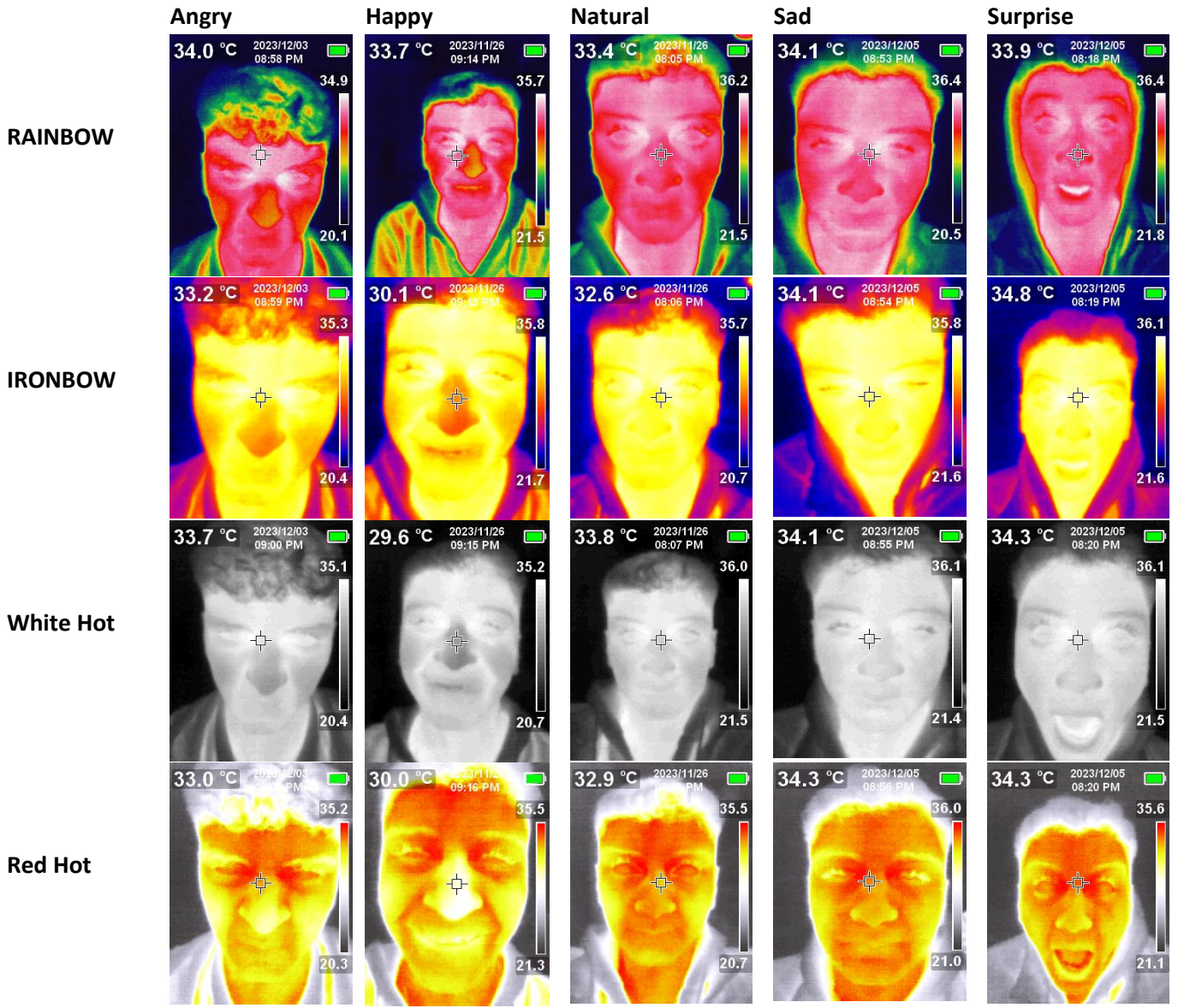

**Fig. 8.** A variety of color palettes are used in our dataset.

ICEBLUE 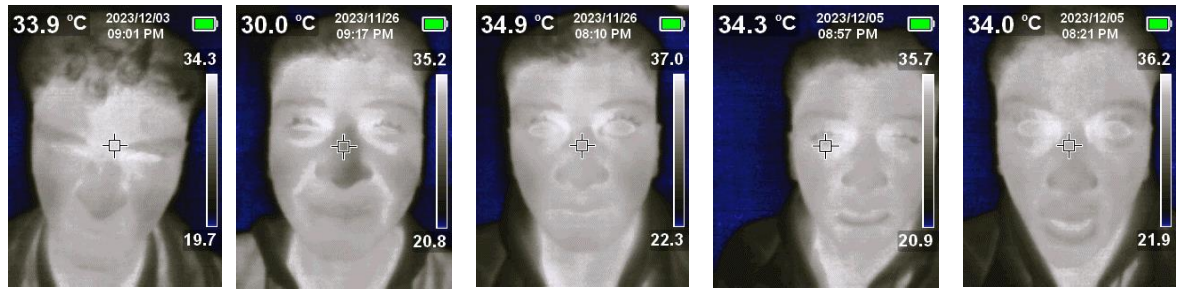

**Fig. 9.** Facial emotion data displays various emotions using different color palettes.

**Thermographic Angles and Varied Zoom Levels**

In compiling the dataset, three distinct Thermographic angles were employed, as illustrated in **Fig .10.** These angles included capturing images from the right side, left side, and front perspectives. Alongside these angles, three distinct zoom levels were meticulously applied. This comprehensive approach ensures a diverse range of perspectives and details within the dataset, enhancing its utility for various analyses and research endeavors.

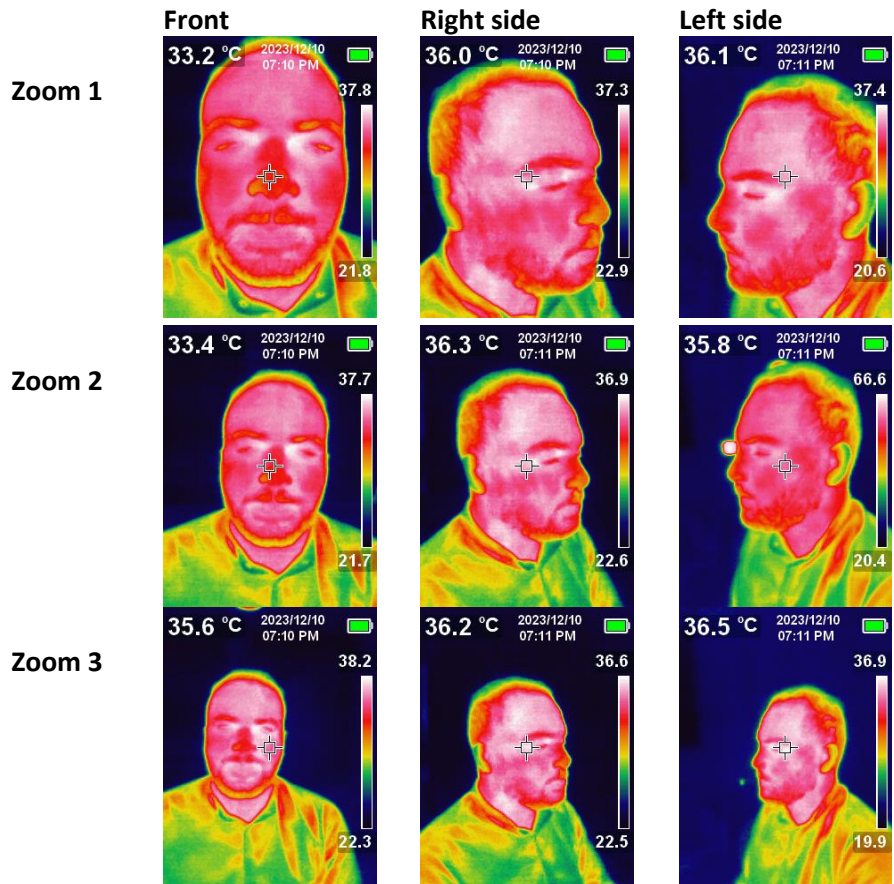

**Fig.10.** Thermographic Data Collection: Angles and Zoom Levels

**SWOT Analysis**

- **Strengths**:

    **Diverse Data Collection**: The dataset comprises **6823** thermal images with a resolution of **19,200** pixels, covering a wide range of temperature from **-10°C to 400°C**. This diversity allows for comprehensive analysis and algorithm development.

    **Facial Emotion Variation**: With **2485** images for five emotions (happy, sad, angry, normal, surprised), the dataset offers a substantial variety for emotion analysis.

    **Color Palette Variability**: Different color palettes such as ICEBLUE, IRONBOW, RAINBOW, Red Hot, and White Hot were used, enhancing the dataset's utility for different applications.

- **Interdisciplinary Utility**: This dataset serves as a valuable resource for a multitude of fields, including Human-Computer Interaction, Computer Vision, Pattern Recognition, and Artificial Intelligence. Researchers and practitioners in these disciplines can utilize the dataset to advance their work in various applications such as facial recognition, emotion analysis, and thermal imaging technology. The dataset's diverse collection of thermal face images, captured under different conditions and using various color palettes, offers a rich source of data for developing and benchmarking algorithms. Additionally, it fosters collaboration between different domains, encouraging innovative approaches to thermal imaging research and applications.
- **Weaknesses**:
  **Limited Sample Size**: While the dataset is comprehensive, with 6823 images, a larger sample size might have improved its generalizability.
  **Dependency on Simulation**: Facial expressions for certain emotions, such as anger, were simulated by volunteers. This may introduce biases or inaccuracies.
  **Sensitivity to Environmental Factors**: Thermal imaging can be affected by environmental conditions such as lighting and background temperature, which may impact the dataset's consistency[15]
- **Opportunities**:
  **Algorithm Development**: Researchers can use this dataset to develop and refine algorithms for facial detection, recognition, and emotion analysis in thermal images.
  **Comparative Studies**: The dataset allows for benchmarking and comparison of different algorithms across various color palettes, offering opportunities for improving accuracy.
  **Facial Motion Recognition**: Insights from this dataset can contribute to advancements in non-verbal communication analysis and emotion detection.
  **Collaborative Research**: The interdisciplinary nature of the dataset encourages collaboration between fields like computer vision, psychology, and neuroscience.
- **Threats**:
  **Data Bias**: Due to the simulated nature of some expressions, the dataset might not fully represent natural facial expressions in thermal imaging.
  Technological Advancements: Rapid advancements in thermal imaging technology may make this dataset outdated in the future.
  In conclusion, the Thermal Facial Dataset offers a robust foundation for research and development in thermal imaging technology, particularly in the areas of facial recognition and emotion analysis. Despite some limitations, its diverse range of images and color palettes provide ample opportunities for algorithm refinement and interdisciplinary collaborations. Continued research and exploration of this dataset can lead to significant advancements in various fields, from security systems to healthcare applications.

## Data Records

Thermal images are crucial for face recognition, facial expression analysis, and face detection due to their ability to capture unique heat patterns, enabling accurate identification and analysis in various conditions, including low light and camouflage [16].

The dataset comprises three primary folders, delineated in **Fig.1**. Each folder contains 2250 thermal images in BMP format: face detection, face recognition, and facial emotion. In total, there are 7750 thermal images across all three folders.

**Facial emotion file description**

Facial emotion data encompasses five primary categories: **538** images depicting happiness, **457** images of sadness, **509** images of anger, **457** images of neutral expressions, and **440** images capturing surprise, totalling **2485** thermal images as shown in **Fig. 2**.

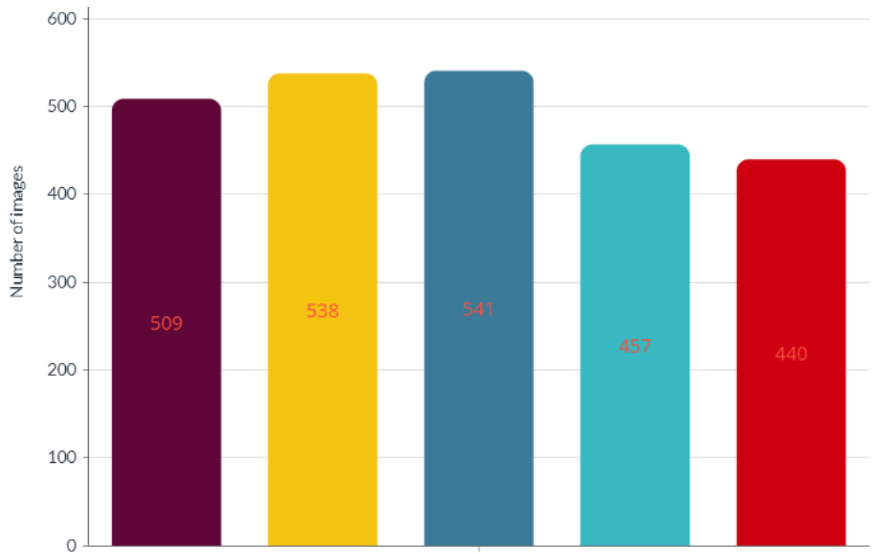

**Fig. 2.** Facial Emotion Data Distribution Across Facial Classes

This dataset provides a comprehensive array of emotions for analysis and study. Within each of these emotion categories, there are various panels featuring different color palettes: RAINBOW, IRONBOW, White Hot, Red Hot, and ICEBLUE, as illustrated in **Fig.3**.

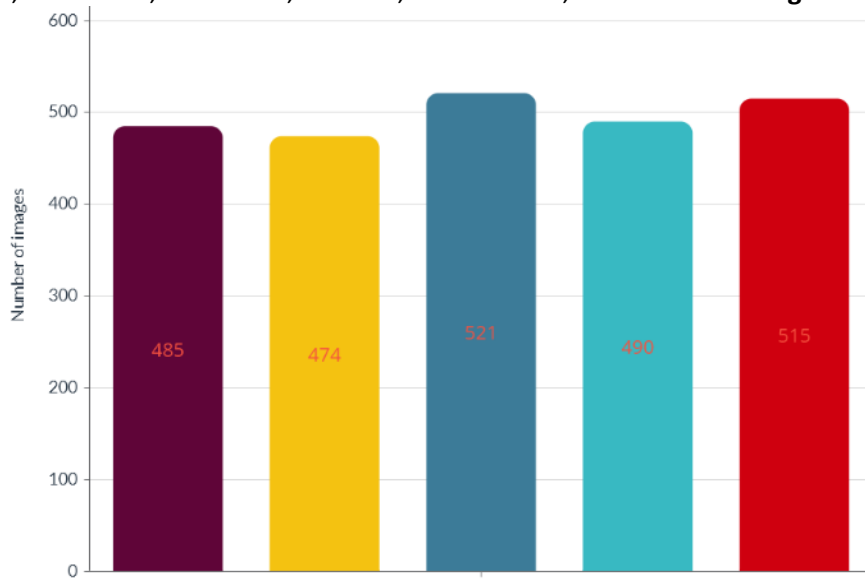

**Fig. 3.** Facial Emotion Data Distribution Across Color Palettes

This diversity in color palettes allows for nuanced analysis and interpretation of thermal imaging data. Moreover, each emotion category includes color dividers with diverse shooting angles and three distinct zoom levels, as indicated in **Fig .4.**

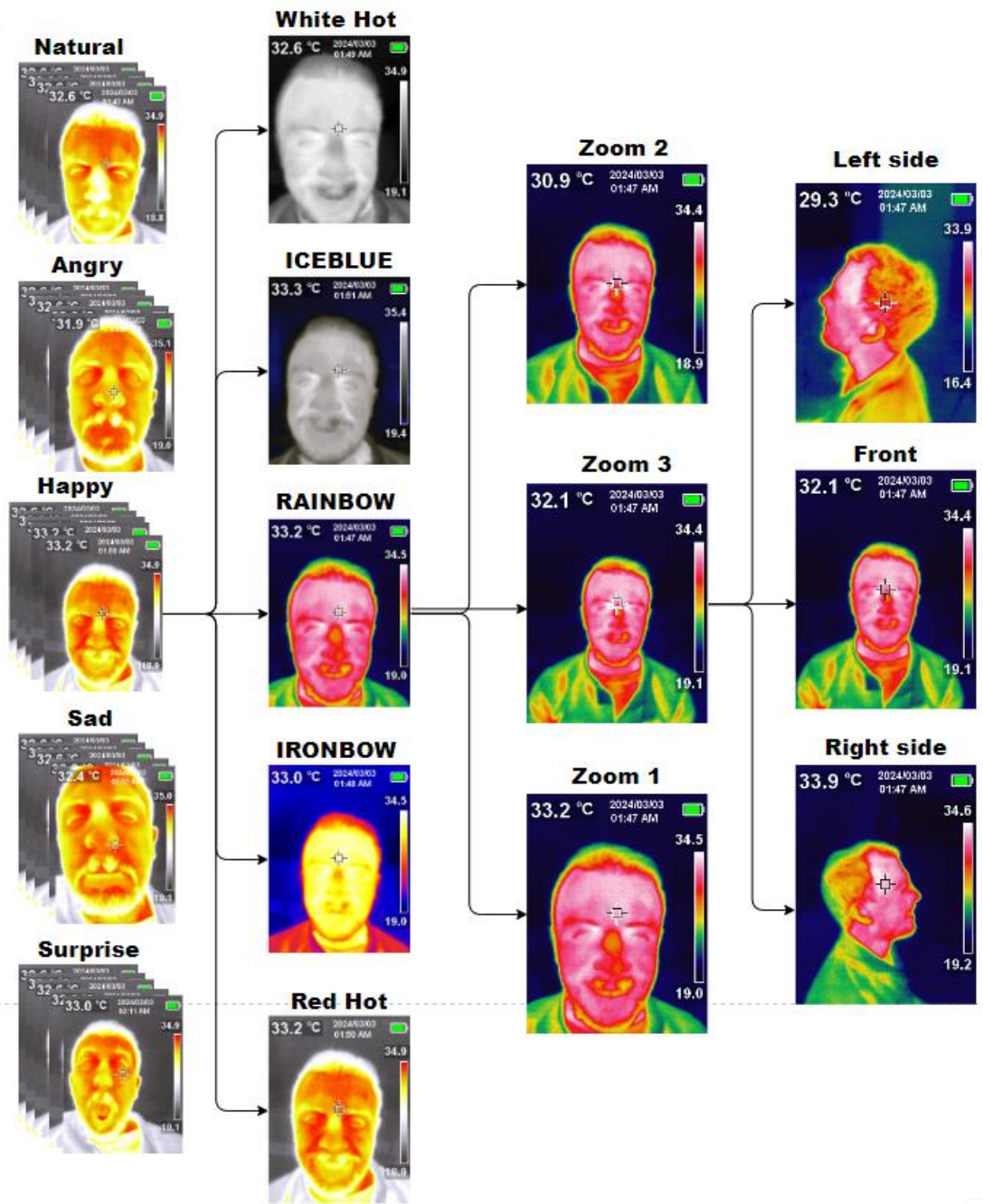

**Fig. 4**. Facial emotion Data structure of the repository

**Face recognition file description**

The face recognition dataset includes ten individuals from three different nationalities, each represented by a number from 1 to 10. **Fig .5** visually represents a sample of these individuals, while **Fig .6** displays the number of images for each person in the dataset.

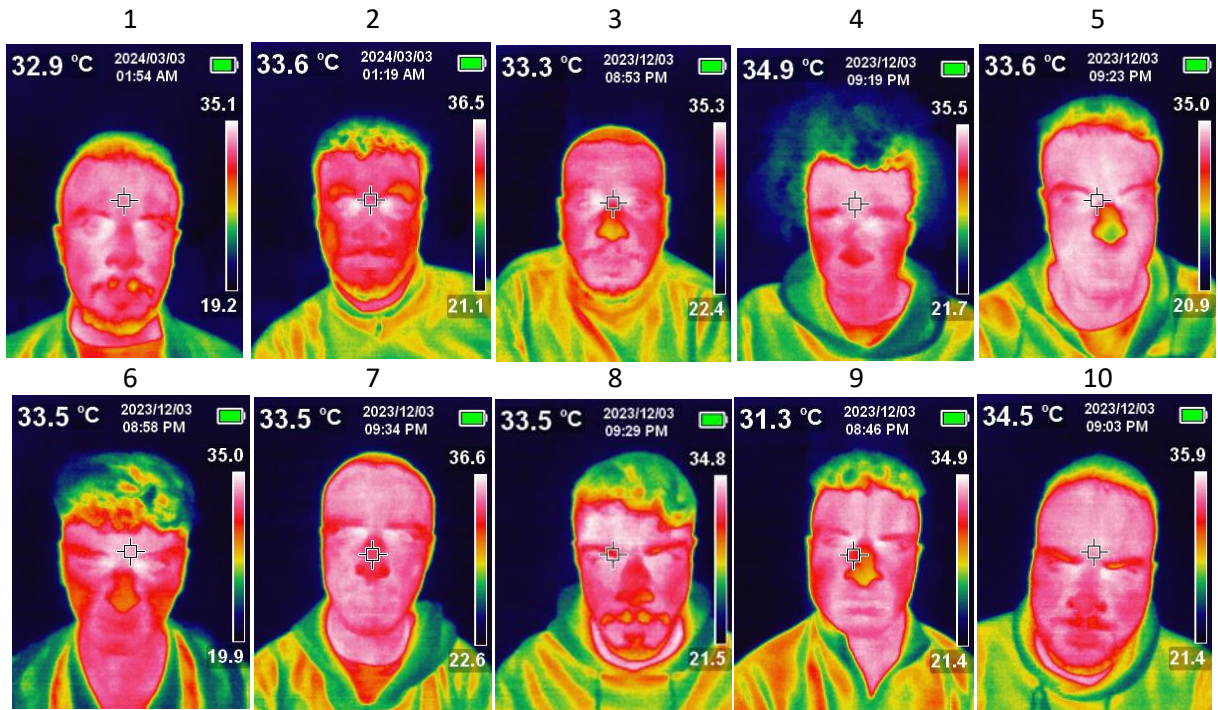

**Fig. 5.** The thermal face recognition dataset visually represents the remaining individuals in the dataset using different people.

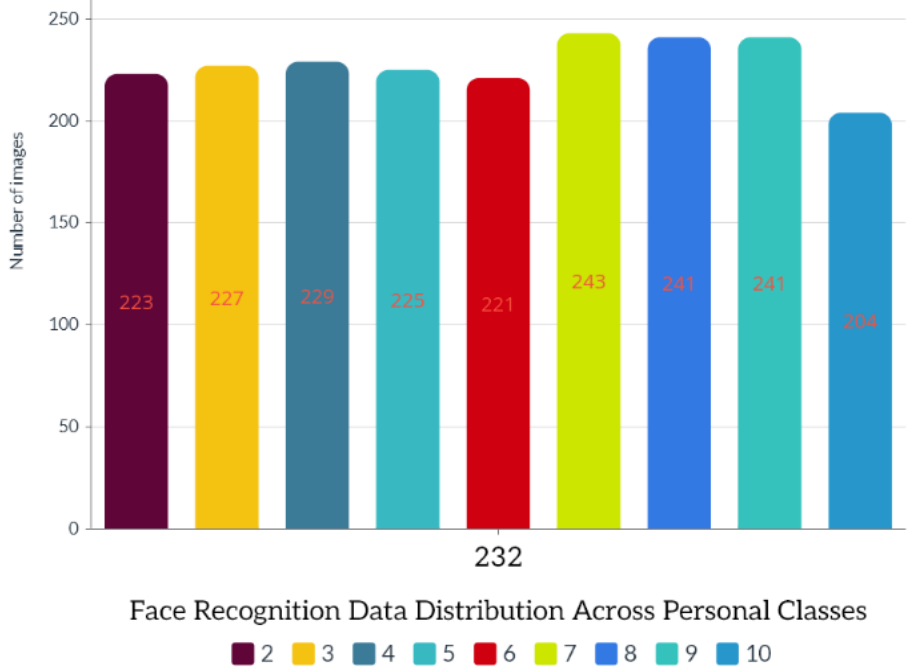

**Fig. 6.** Face Recognition Data Distribution Across Personal Classes

**Face detection file description**

The face detection dataset comprises **2284** images and encompasses all data, including facial emotions and facial recognition, it is categorized into different color palettes, each with varying shooting angles, and each angle is captured at different zoom levels, as illustrated in **Fig .7**.

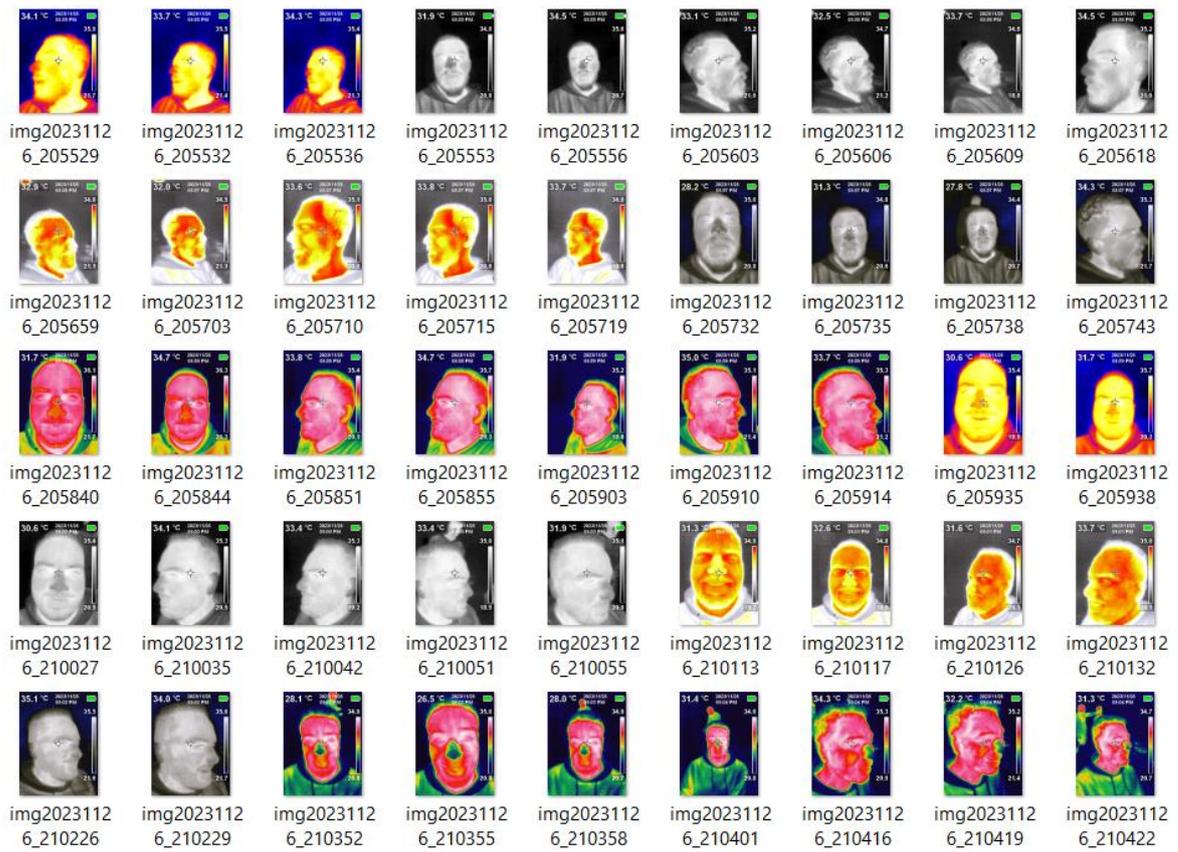

**Fig. 7.** Face detection data displays

**Fig .8** shows the distribution of face detection numbers across different color palettes: RAINBOW, IRONBOW, White Hot, Red Hot, and ICEBLUE. This provides a quick reference for researchers to understand how the dataset is categorized based on color options, aiding in the selection of appropriate color schemes for analysis or algorithm development. Overall, **Fig .8** offers valuable insights into the organization of the dataset by color palettes.

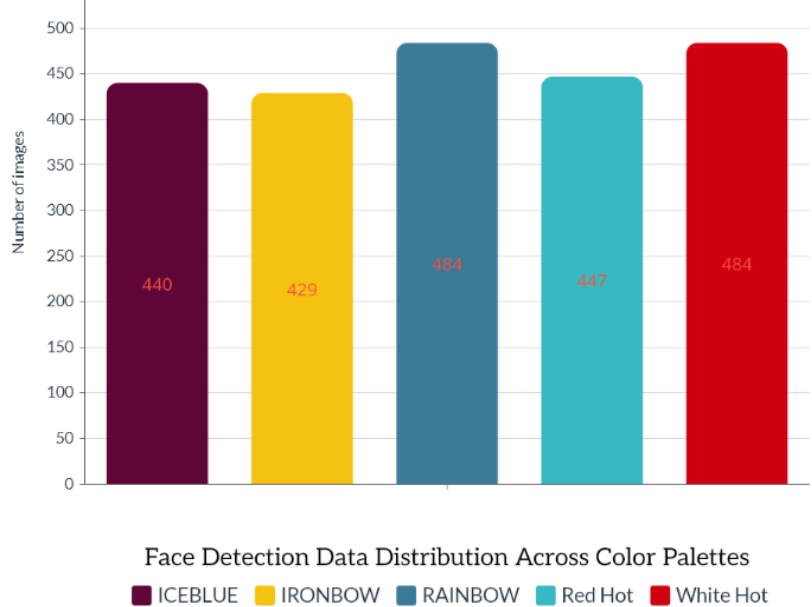

**Fig. 8.** Face Detection Distribution Across Color Palettes

# Technical Validation

**LIMITATIONS**

During the facial expression recognition data collection process for this study, several challenges and limitations were encountered. Unlike visual images, which rely solely on facial features, thermal imaging introduces a correlation between facial expression and temperature, influenced by psychological states. To address this, specific measures were implemented during data collection. For instance, with the angry group, volunteers were instructed to simulate anger by maintaining specific facial expressions such as forceful eye-opening and clenching their jaws. These efforts were made to create thermal images that could accurately represent emotional states. Additionally, different color palettes were used to compile the images, addressing the challenges posed by red radiation rays produced in temperature images from the volunteers' faces.

Despite these challenges, the dataset provides valuable insights into the application of thermal imaging for facial recognition, particularly in the realm of facial expression recognition. These insights contribute to the ongoing advancement of thermal imaging technology, offering the potential for improved understanding and interpretation of emotional states through facial features and temperature correlations. The integration of thermal imaging with facial recognition opens new avenues for studying emotional responses and enhancing understanding in various fields such as psychology, healthcare, and security.

# Code Availability

Custom code was not utilized for this study. The versions and parameters of the bioinformatics tools employed in this research were delineated in the Methods section. Any instance where a parameter was utilized with a value other than its default setting was explicitly mentioned earlier.

# Author contributions

All authors contributed equally to this work.

# Competing interests

The authors declare that there are no competing interests conflicting with this research